\documentclass[letterpaper,times]{sae}
\definecolor{SAEblue}{RGB}{1,160,233}

\usepackage[justification=justified, singlelinecheck=false]{caption}  
\DeclareCaptionFont{eightpt}{\fontsize{8pt}{11pt}\selectfont #1}
\usepackage{lineno}
\usepackage[utf8]{inputenc}

\usepackage{array}
\newcolumntype{L}[1]{>{\raggedright\let\newline\\\arraybackslash\hspace{0pt}}p{#1}}
\newcolumntype{C}[1]{>{\centering\let\newline\\\arraybackslash\hspace{0pt}}p{#1}}
\newcolumntype{R}[1]{>{\raggedleft\let\newline\\\arraybackslash\hspace{0pt}}p{#1}}

\usepackage{graphicx}
\usepackage{multirow}
\usepackage{amsmath}

\newcommand{\ignore}[1]{}

\usepackage{amsmath} 
\usepackage{amssymb} 
\usepackage{cite}
\usepackage{nameref}

\usepackage{graphicx}
\usepackage{tikz}
\usepackage{pgfplots}
\usepackage{subcaption}
\usepackage{makecell}  
\usepackage[nolist,nohyperlinks]{acronym}
\usepackage{booktabs}
\usepackage{enumitem}
\usepackage[bottom]{footmisc}  
\usepackage{mathtools}

\makeatletter
\def\@seccntformat#1{%
  \expandafter\csname c@#1\endcsname\c@section
  }
\let\NAT@parse\undefined
\makeatother
\usepackage[hidelinks]{hyperref}  

\PaperTitle{FSOCO: The Formula Student Objects in Context Dataset}
\usepackage{calc} 

\makeatletter 
\renewcommand\@biblabel[1]{#1. } 
\makeatother

\AddAuthor{Niclas Vödisch*}{ETH Zürich, Switzerland}
\AddAuthor{David Dodel*}{LMU Munich, Germany}
\AddAuthor{Michael Schötz*}{Deggendorf Institute of Technology, Germany}
\AddNote{* All authors contributed equally to this work. \\ Contact: {\tt fsoco.dataset@gmail.com}}
\AddHeader{Published in: \textit{SAE Int. J. of CAV} 5(1):2022, doi:10.4271/12-05-01-0003}

\pgfplotsset{compat=1.12}

\newcommand{\red}[1]{{#1}}

\newcommand{\reffig}{Figure~\ref}
\newcommand{\refsec}[1]{\red{the \nameref{#1} section}}
\newcommand{\reftab}{Table~\ref}
\newcommand{\refeqn}{Equation~\ref}

\begin{acronym}
\acro{FSD}{Formula Student Driverless}
\acro{FSG}{Formula Student Germany}
\acro{FSOCO}{Formula Student Objects in Context}
\acro{AP}{average precision}
\acro{IoU}{Intersection over Union}
\acro{OOD}{out-of-distribution}
\acro{CLI}{command-line interface}
\end{acronym}

\usepackage[edges]{forest}
\definecolor{folderbg}{RGB}{124,166,198}
\definecolor{folderborder}{RGB}{110,144,169}
\newlength\Size
\tikzset{%
  folder/.pic={%
    \filldraw [draw=folderborder, top color=folderbg!50, bottom color=folderbg] (-1.05*\Size,0.2\Size+5pt) rectangle ++(.75*\Size,-0.2\Size-5pt);
    \filldraw [draw=folderborder, top color=folderbg!50, bottom color=folderbg] (-1.15*\Size,-\Size) rectangle (1.15*\Size,\Size);
  },
  file/.pic={%
    \filldraw [draw=folderborder, top color=folderbg!5, bottom color=folderbg!10] (-\Size,.4*\Size+5pt) coordinate (a) |- (\Size,-1.2*\Size) coordinate (b) -- ++(0,1.6*\Size) coordinate (c) -- ++(-5pt,5pt) coordinate (d) -- cycle (d) |- (c) ;
  },
}
\forestset{%
  declare autowrapped toks={pic me}{},
  pic dir tree/.style={%
    for tree={%
      folder,
      font=\ttfamily,
      grow'=0,
    },
    before typesetting nodes={%
      for tree={%
        edge label+/.option={pic me},
      },
    },
  },
  pic me set/.code n args=2{%
    \forestset{%
      #1/.style={%
        inner xsep=2\Size,
        pic me={pic {#2}},
      }
    }
  },
  pic me set={directory}{folder},
  pic me set={file}{file},
}

\begin{document}
\maketitle

\thispagestyle{empty}
\pagestyle{empty}

\begin{abstract}

This paper presents the FSOCO dataset, a collaborative dataset for vision-based cone detection systems in Formula Student Driverless competitions. It contains human annotated ground truth labels for both bounding boxes and instance-wise segmentation masks. The data buy-in philosophy of FSOCO asks student teams to contribute to the database first before being granted access ensuring continuous growth. By providing clear labeling guidelines and tools for a sophisticated raw image selection, new annotations are guaranteed to meet the desired quality. The effectiveness of the approach is shown by comparing prediction results of a network trained on FSOCO and its unregulated predecessor. The FSOCO dataset can be found at \texttt{\href{https://fsoco.github.io/fsoco-dataset/}{fsoco.github.io/fsoco-dataset/}}.

\end{abstract}

\section{Introduction}
\label{sec:introduction}

\begin{figure}[t]
    \centering
    \begin{subfigure}{\linewidth}
        \centering
        \includegraphics[width=\linewidth]{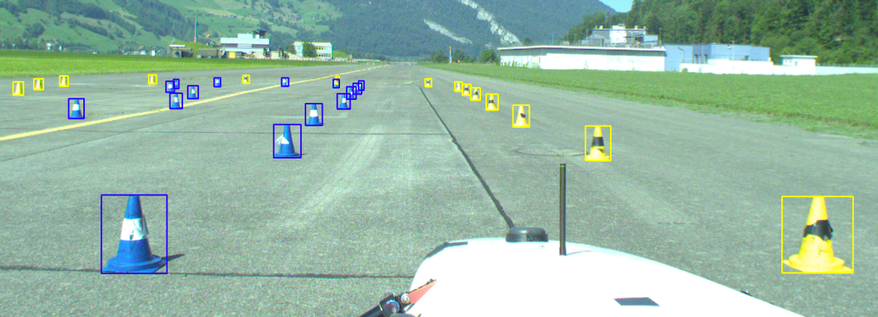}
    \end{subfigure}
    \\[1ex]
    \begin{subfigure}{\linewidth}
        \centering
        \includegraphics[width=\linewidth]{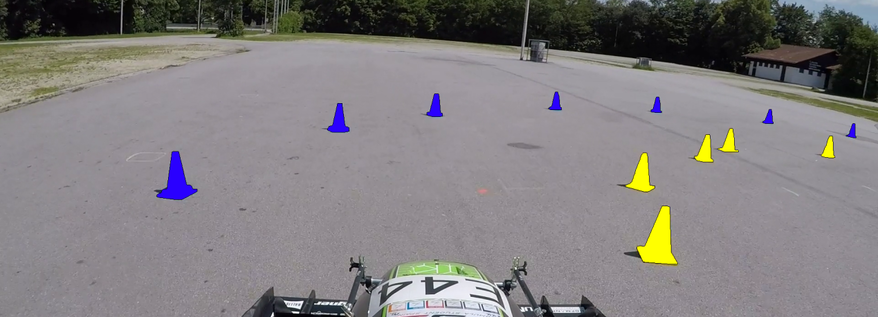}
    \end{subfigure}
    \caption{The FSOCO dataset contains human annotated bounding boxes and instance-wise segmentation masks for traffic cones as being used in the Formula Student Driverless competitions.}
    \label{fig:cover_figure}
\end{figure}

Visual scene interpretation and object detection are key requirements for autonomous vehicles. This holds true for both self-driving passenger cars and trucks as well as for autonomous race cars, which have attracted more and more attention in recent years due to competitions such as Roborace or \ac{FSD}. In contrast to efforts made towards autonomous urban driving, these competitions provide a fixed set of rules and a description of the expected scenarios. In particular, \ac{FSD} differs from Roborace by a stronger focus on perception, as the vehicle must be able to navigate the track without a pre-determined map. Whereas in Roborace, the software architecture \cite{Betz2019} and main development focus~\cite{Caporale2019} are put on control systems due to the previously known map. In \ac{FSD}, the race track is marked by cones allowing the race cars to complete laps without human supervision. The \ac{FSOCO} dataset contains a large set of images with human annotations of these cones. As shown in \reffig{fig:cover_figure}, label types include both bounding boxes and instance-wise segmentation masks. Access is granted based on a contribution-based policy, i.e., student teams are required to provide new annotations before being able to download the existing database. This procedure ensures the continuous growth of the dataset as well as a label quality up to the user's needs.

\subsection{Formula Student Driverless}
Formula Student is a series of competitions, in which teams of students compete against each other for the best design and construction of an open-cockpit race car. The competitions are split into static and dynamic disciplines, the former focusing mainly on the design while the latter focusing on the build in the form of time trial races.

The competition series started with internal combustion race cars with FS Michigan in 1981 and has since expanded internationally. National competitions are held in 19 countries currently and started to include Electrical and Driverless vehicles from 2010 and 2017 on, respectively.
The largest competition with respect to the number of participants is \ac{FSG}, with the latest edition in 2019 gathering 3,500 participants and 119 teams from 24 countries\cite{fsg_press19, fsg_teams19}.
It was also at \ac{FSG} that the Electrical and Driverless categories were first introduced.

\textbf{FSD Dynamic Disciplines}
The dynamic disciplines are time-trials on different track layouts in which the performance of all cars is measured relative to the best performing car. The layout, procedure, and scoring of the races are defined in the event rules, serving as a guideline for the cars' design. \textit{Skid Pad} is a fixed track layout race consisting of two overlapping concentric circles with a straight segment going through the middle. The schematic of the track resembles an infinity symbol with a line going through its middle portion. \textit{Acceleration} is another fixed track layout race that in turn consists of a straight track. Both \textit{Autocross} and \textit{Track Drive} use a circuit track with no fixed layout other than restrictions regarding ranges of curvatures and lengths of track segments that should be expected, e.g, minimum curvature of curves, maximal length of straights, and approximate total track length.

The disciplines can roughly be split into the fixed layout and variable layout categories with respect to the available track information from the competition handbook. With a fixed layout, \textit{Skid Pad} and \textit{Acceleration} put an emphasis on the base vehicle's performance and allow the autonomous system to benefit from reliable prior knowledge. The variable layout tracks, on the other hand, require the cars to find the optimal racing line on-the-fly, possibly leveraging information from previous runs.

\begin{figure}[b]
    \vspace{2ex}
    \centering
    \captionsetup[subfigure]{justification=centering}
    \begin{subfigure}[t]{0.24\linewidth}
        \centering
        \includegraphics[width=\linewidth]{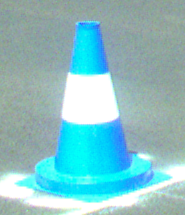}
        \caption{blue}
    \end{subfigure}
    \begin{subfigure}[t]{0.24\linewidth}
        \centering
        \includegraphics[width=\linewidth]{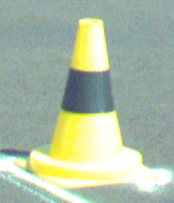}
        \caption{yellow}
    \end{subfigure}
    \begin{subfigure}[t]{0.24\linewidth}
        \centering
        \includegraphics[width=\linewidth]{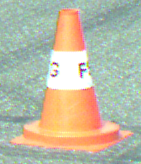}
        \caption{small orange}
    \end{subfigure}
    \begin{subfigure}[t]{0.24\linewidth}
        \centering
        \includegraphics[width=\linewidth]{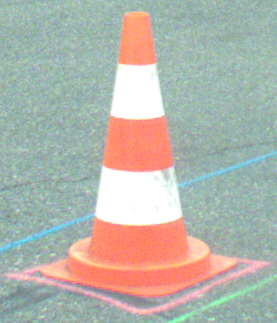}
        \caption{large orange}
    \end{subfigure}
    \caption{\ac{FSOCO} supports five object classes. The four main classes are shown here, the fifth class \textit{other} includes all cones that are not rules compliant.}
    \label{fig:object_classes}
\end{figure}

Common among all tracks is the use of traffic cones of different colors as the only artifact used to mark track limits. The shape, dimensions, and exact model of the traffic cones are defined in the competition handbook allowing teams to test in competition-like settings. Moreover, the color of the cones is relevant. Blue and yellow cones delimit the left and right sides of the track, respectively. Large orange cones signal the start and finish lines, while small orange cones are used on both sides for entry and exit lanes. \reffig{fig:object_classes} shows the different cones.

\subsection{FSD Autonomous Systems}
\label{sub:fsd_as}
The task solved by the cars in the dynamic disciplines can thus be boiled down to perception of the cones, localization, planning, and finally control. This scheme closely resembles the Sense-Plan-Act robotic paradigm and is usually implemented as a Hierarchical Planner~\cite{Arkin1998}. Note that no prediction of the environment's state changes is necessary since the objects in the environment of the dynamic disciplines are static, except when hit by the car. Additionally, even in cases in which a car hits cones, prediction is often not possible since the cone will be in the sensor blind spot, either stuck under or on the car's bodywork. On a system-level this lack of necessity for environmental prediction is the biggest conceptual difference to urban autonomous vehicle systems, while on a component-level the small number of distinct object classes and scene backgrounds simplifies the perception task.

We will now focus on implementations of the perception system.
The most common environmental sensors employed in \ac{FSD} race cars are RGB cameras in both monocular and stereo configurations \cite{zeilinger2017design, Valls2018, Gosala2018, Andresen2020, Strobel2020}.
Based on the dense, high-dimensional output of these sensors, object detection methods are employed to fit the cones to geometries on the image plane that are processed further downstream. For this, most teams rely on deep learning single-shot detectors like YOLOv2 \cite{Dhall2019} and YOLOv3 \cite{Andresen2020, Strobel2020}.
To the best of our knowledge, no team has used pixel-based methods as semantic and instance segmentation to this end yet. Thus, we hope to provide a base dataset with FSOCO.
Less widespread are approaches that use classical object detection methods like ones based on the Hough transform \cite{starkstrom_academy18}.
Finally, combining monocular or stereo cameras with LiDARs is the most common setup, e.g., used in \cite{Valls2018, Gosala2018, Andresen2020}.
By combining the different modalities through sensor fusion, redundancy and greater precision can be achieved as the sensors have strengths and weaknesses that complement each other.

\subsection{Related Datasets}

In the last decade many datasets, providing ground truth labels for image-based object detection, have been published. Some of them cover a wide range of object classes, e.g., ImageNet~\cite{Deng2009} and Microsoft COCO~\cite{Lin2014}, while others focus on a specific use case.

In the field of autonomous driving, KITTI~\cite{Geiger2013} was one of the pioneering datasets including not only annotated camera images but LiDAR point clouds, GPS/IMU data, and calibration parameters of all sensors. By providing an easy-to-use evaluation server for an unpublished test set, KITTI accelerated the development of perception and tracking algorithms for self-driving vehicles. Cityscapes~\cite{Cordts2016} is another notable large-scale dataset focusing on semantic instance-level scene labeling from 50 different cities. Following the latest requirements of the research community, the dataset has been extended by 3D bounding boxes~\cite{gaehlert2020}. Recent multi-modal datasets such as Argoverse~\cite{chang2019} and nuScenes~\cite{caesar2020} include pre-built HD maps fostering the usage of additional offline route information.

The generation of publicly available datasets requires the consideration of several aspects. COCO~\cite{Lin2014} used workers on Amazon's Mechanical Turk to crowdsource annotation tasks. In order to ensure that workers understood the necessary accuracy of the segmentation masks, they have to successfully finish a training task that compared their segmentation with approved ground truth. Additionally, each submitted image was reviewed by multiple other workers to judge its quality. Poorly annotated images were discarded and sent back to the set of raw images. Despite also using Mechanical Turk, ObjectNet~\cite{barbu2019} focused more on the data collection process and aiding the annotators. Workers were asked to upload images of objects in their homes while providing detailed instructions on viewpoints and backgrounds. Objects that pose a safety concern or violate people's privacy were rejected. For the same reason, Argoverse~\cite{chang2019} blurred license plates and faces.

In the context of Formula Student, the first attempt towards a collaborative dataset was made by FSOCO legacy\footnote{We use \textit{legacy} to refer to the discontinued first version of FSOCO.}, that also worked on a data buy-in basis with a minimal contribution of 600 images. However, no official guidelines regarding the labeling were posed except for rough documentation by each contributor. The legacy version demonstrated the demand for such a collaborative dataset from the \ac{FSD} community reaching 39 contributing teams, 44,195 images, and 316,669 labeled cones within one year of existence. Unfortunately, due to very large differences in the quality and creation of the constituent team datasets, only a fraction of the whole dataset was usable after very time-intensive manual data pre-processing. While other datasets, such as BDD100K~\cite{Yu2020} or nuScenes~\cite{caesar2020}, contain traffic cones as well, the captured environment differs largely from \ac{FSD} race tracks and the available datasets do not differentiate between cone colors.

\section{The Dataset}
\label{sec:the_dataset}
The \ac{FSOCO} dataset is a semi-open, data buy-in dataset for vision-based perception tasks in \ac{FSD} competitions. It is motivated by three guiding ideas: 1) high quality, 2) collaboration, and 3) easy access.

First, the quality of data and labels contained in the dataset is assured by the steps taken during the data collection, see \refsec{sec:data_collection}, and, in particular, a partially automated contribution procedure designed to ensure that the labeling guidelines are correctly followed.
Second, the collaborative nature of the dataset stems from its unique data buy-in contribution procedure. To ensure the organic growth of the dataset, access to its current training set is only granted to parties that have contributed to it. Additionally, teams with excess raw data can donate it to be labeled by teams that do not have access to the necessary hardware, e.g., cones, cameras, and an automotive platform.
Finally, easy access to the dataset for all interested parties is guaranteed by keeping all the tools and documentation needed to participate free of charge. Moreover, no prior knowledge is expected or needed to be able to understand the contribution guidelines.

\begin{figure}[t]
    \centering
    \captionsetup[subfigure]{justification=centering}
    \begin{subfigure}[t]{0.24\linewidth}
        \centering
        \includegraphics[width=\linewidth]{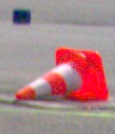}
        \caption{knocked over}
    \end{subfigure}
    \begin{subfigure}[t]{0.24\linewidth}
        \centering
        \includegraphics[width=\linewidth]{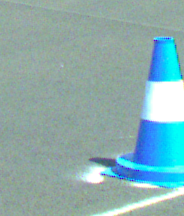}
        \caption{truncated}
    \end{subfigure}
    \begin{subfigure}[t]{0.24\linewidth}
        \centering
        \includegraphics[width=\linewidth]{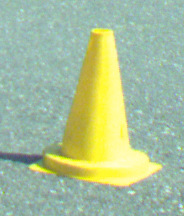}
        \caption{tape removed}
    \end{subfigure}
    \begin{subfigure}[t]{0.24\linewidth}
        \centering
        \includegraphics[width=\linewidth]{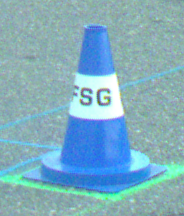}
        \caption{event sticker}
    \end{subfigure}
    \caption{Cones can be annotated by three different object tags, where (c) and (d) are combined in a single tag.}
    \label{fig:object_tags}
\end{figure}

\subsection{Annotations}
\label{ssec:annotations}
\begin{figure}[b]
    \vspace{2ex}
    \centering
    \captionsetup[subfigure]{justification=centering}
    \begin{subfigure}{\linewidth}
        \centering
        \includegraphics[width=\linewidth]{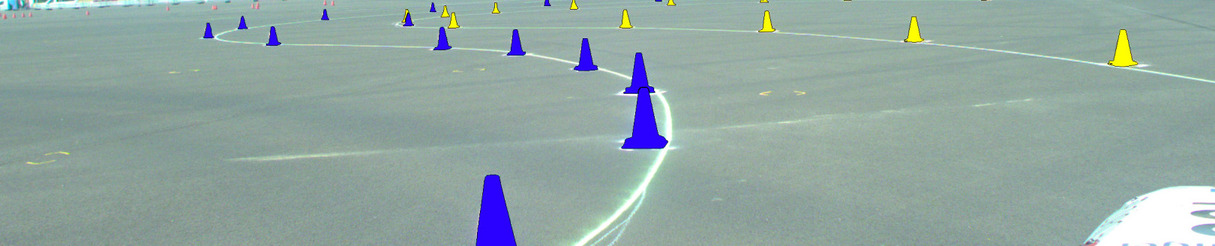}
    \end{subfigure}
    \\[1ex]
    \begin{subfigure}{\linewidth}
        \centering
        \includegraphics[width=\linewidth]{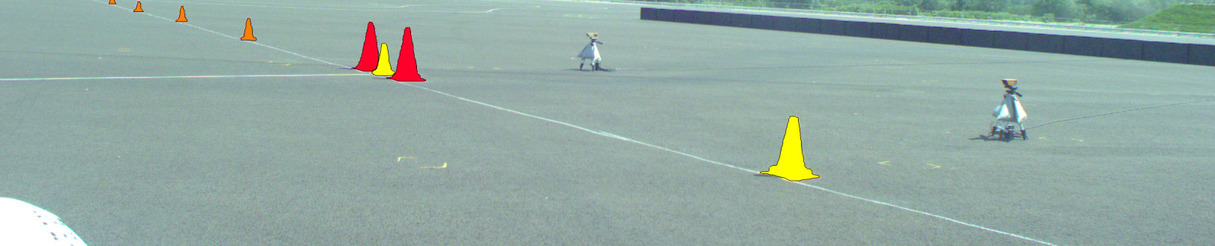}
    \end{subfigure}
    \caption{Example images showing instance-wise segmentation masks.}
    \label{fig:segmentation_samples}
\end{figure}

The \ac{FSOCO} dataset contains two different types of labels: 1) bounding boxes and 2) instance-wise segmentation masks. Example annotations are shown in \reffig{fig:bounding_box_samples} and \reffig{fig:segmentation_samples}, respectively.

We distinguish five different object classes corresponding to the four official cones, i.e., blue, yellow, small orange, and large orange, as well as an additional class to account for cones that are not compliant with the \ac{FSG} rules. We decided to include such an auxiliary class to provide an option to our contributors to re-utilize their previously labeled data without having to delete many objects. However, we do not consider those cones in the contribution requirements and our data analysis. Examples of the four main object classes are given in \reffig{fig:object_classes}.

Each of the cones can be further annotated with three different object tags as shown in \reffig{fig:object_tags}: 1) \textit{knocked over}, 2) \textit{truncated}, and 3) \textit{tape removed or event sticker}. Cones can be knocked over due to various reasons, e.g., to purposefully highlight the border of the dynamic area or because they have been hit by a car. We ask our contributors to tag cones as truncated if they are either cut by the image border or occluded by another cone. The third tag indicates that the cone's appearance has been altered including the (partial) removal of the tape and the attachment of stickers.

\subsection{Data Format}
\label{ssec:data_format}
\begin{figure}[t]
\centering
\includegraphics[width=.6\linewidth]{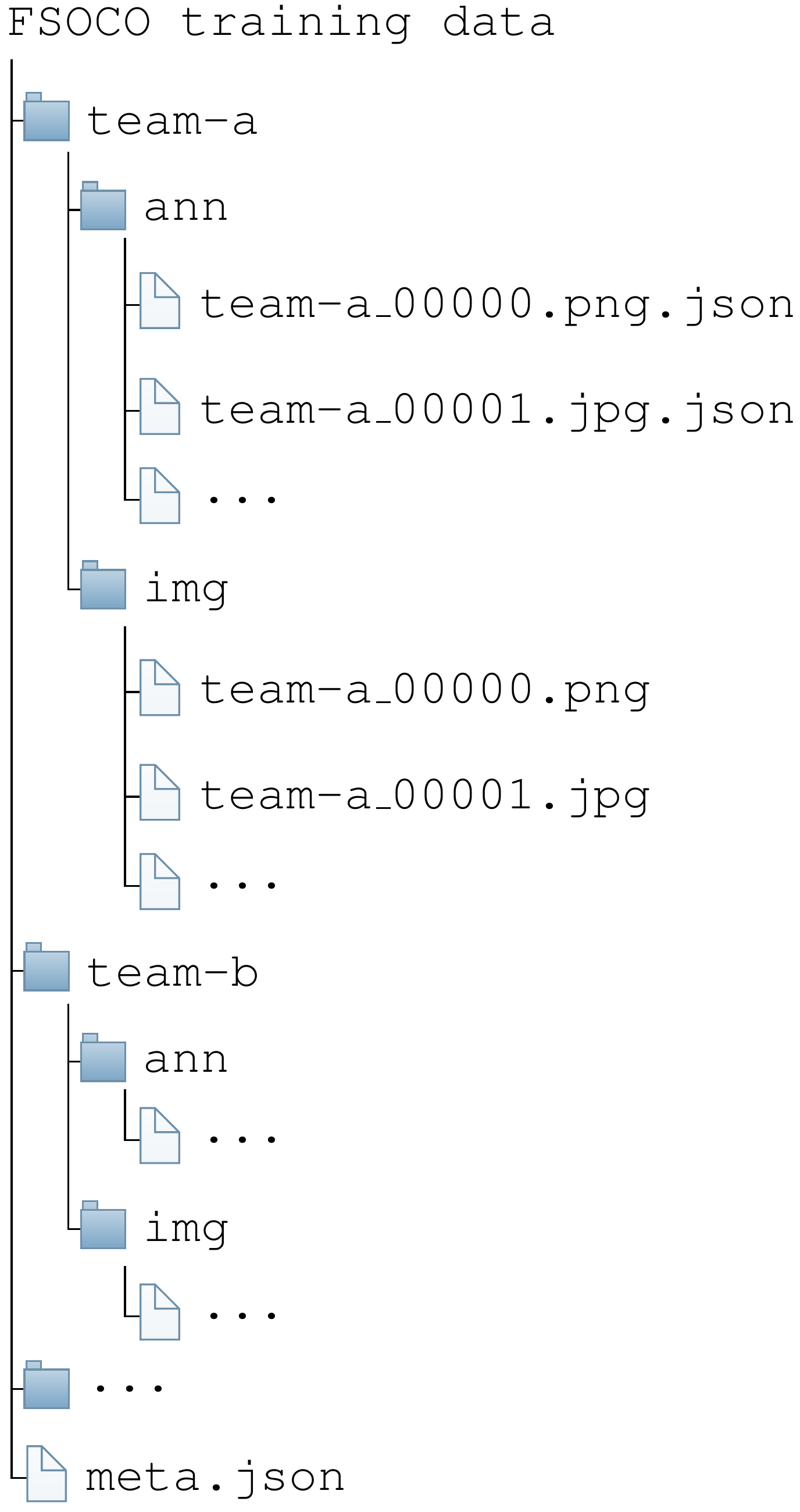}
\caption{Directory layout of the FSOCO dataset.}
\label{fig:dir_layout}
\end{figure}

\reffig{fig:dir_layout} shows the directory layout of the FSOCO dataset. We follow the data structure and label format of Supervisely\footnote{\tt\small\href{https://www.supervise.ly}{supervise.ly}}, an online labeling platform, but also provide several label converters to simplify usage of our data, see \refsec{ssec:label_converters}.

Each team's contribution, independent of whether they labeled their own images or raw data donated by another team, can be found in a subfolder named with a unique team identifier to give them the appropriate credits. A subfolder comprises two directories, containing the images and the corresponding annotation files, respectively. Image and label files follow the structure \texttt{<team-ID>\_<5-digit-number>.<suffix>}, where the suffix is an image format, followed by \texttt{.json} for labels. Supported image extensions include \texttt{png}, \texttt{jpg}, and \texttt{jpeg} using either uppercase or lowercase spelling.

Unlike the example of \texttt{team-a} in \reffig{fig:dir_layout}, image and annotation files do not always have the same team identifier as the corresponding folder. Generally, filenames indicate the source of the raw data to allow for recovering image sequences across multiple team subdirectories.

\subsection{Statistical Analysis}
\label{ssec:statistical_analysis}
In the following we present relevant qualitative and quantitative statistics with respect to the dataset's state at the time of submission. For the most recent numbers we refer the reader to our website.
\ac{FSOCO} currently contains raw data from four different teams with varying sensor setups. Moreover, data from different sensor setups is available for each team. The on-board sensor setups differ in their scene perspective, although most being similar to the view of a front-facing camera mounted at the height of the rear-view mirror of a regular car.

\begin{figure}[t]
    \centering
    \captionsetup[subfigure]{justification=centering}
    \begin{subfigure}{0.495\linewidth}
        \centering
        \includegraphics[width=\linewidth]{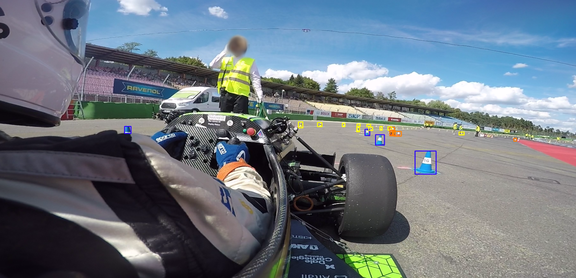}
        \caption{sunny}
        \label{sfig:sunny}
    \end{subfigure}
    \begin{subfigure}{0.495\linewidth}
        \centering
        \includegraphics[width=\linewidth]{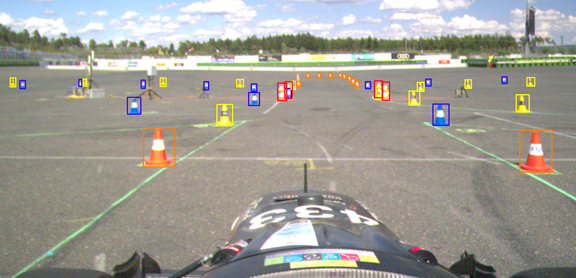}
        \caption{cloudy}
        \label{sfig:cloudy}
    \end{subfigure}
    \\[1ex]
    \begin{subfigure}{0.495\linewidth}
        \centering
        \includegraphics[width=\linewidth]{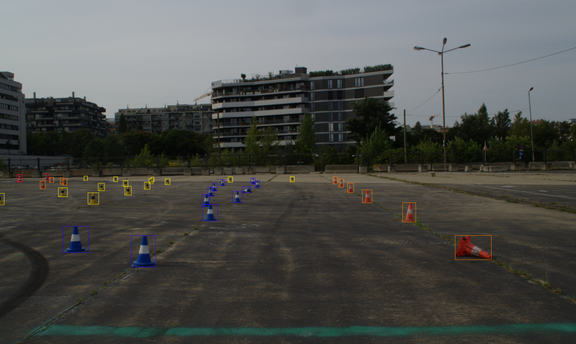}
        \caption{dim light}
        \label{sfig:dim-light}
    \end{subfigure}
    \begin{subfigure}{0.495\linewidth}
        \centering
        \includegraphics[width=\linewidth]{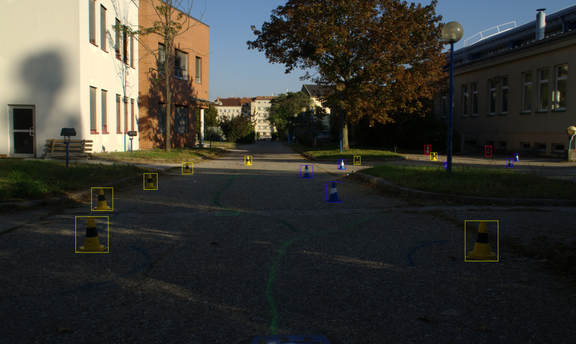}
        \caption{drop shadows}
        \label{sfig:drop-shadows}
    \end{subfigure}
    \\[1ex]
    \begin{subfigure}{0.495\linewidth}
        \centering
        \includegraphics[width=\linewidth]{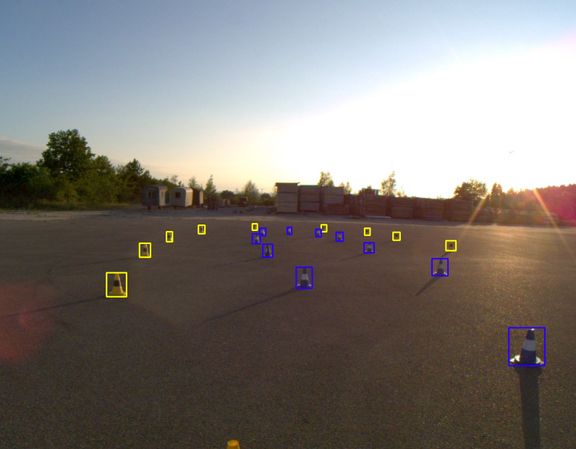}
        \caption{sunset}
        \label{sfig:sunset}
    \end{subfigure}
    \begin{subfigure}{0.495\linewidth}
        \centering
        \includegraphics[width=\linewidth]{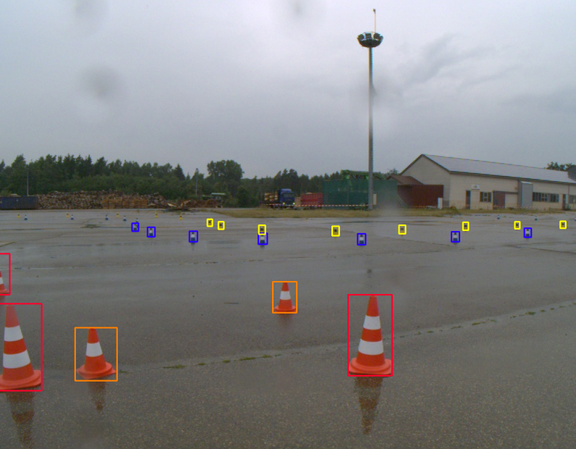}
        \caption{rainfall}
        \label{sfig:rainfall}
    \end{subfigure}
    \caption{Example images showing bounding boxes for different lighting and weather conditions.}
    \label{fig:bounding_box_samples}
\end{figure}

The dataset comprises several different lighting and weather conditions, covering a large number of possible settings. As can be seen in \reffig{fig:bounding_box_samples}, lighting conditions from sunny to cloudy and dim light are present (\reffig{sfig:sunny}-\ref{sfig:dim-light}). Additionally, hard varying conditions as drop shadows are also included (\reffig{sfig:drop-shadows}). Even more difficult weather and lighting conditions are sunset, leading to overexposure, and rainfall with water droplets adhering to the lens, causing deformation and occlusion (\reffig{sfig:sunset},~\ref{sfig:rainfall}).

\begin{figure*}[t]
    \centering
    \captionsetup[subfigure]{justification=centering}
    \begin{subfigure}{0.495\linewidth}
        \centering
        \includegraphics[width=\linewidth]{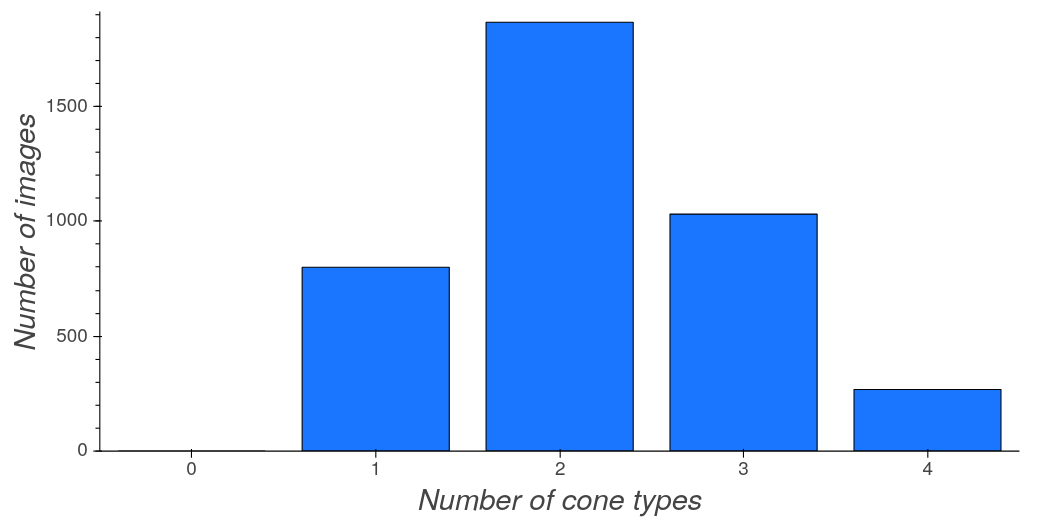}
        \caption{Distinct object classes per image.}
        \label{sfig:classes-per-img}
    \end{subfigure}
    \begin{subfigure}{0.495\linewidth}
        \centering
        \includegraphics[width=\linewidth]{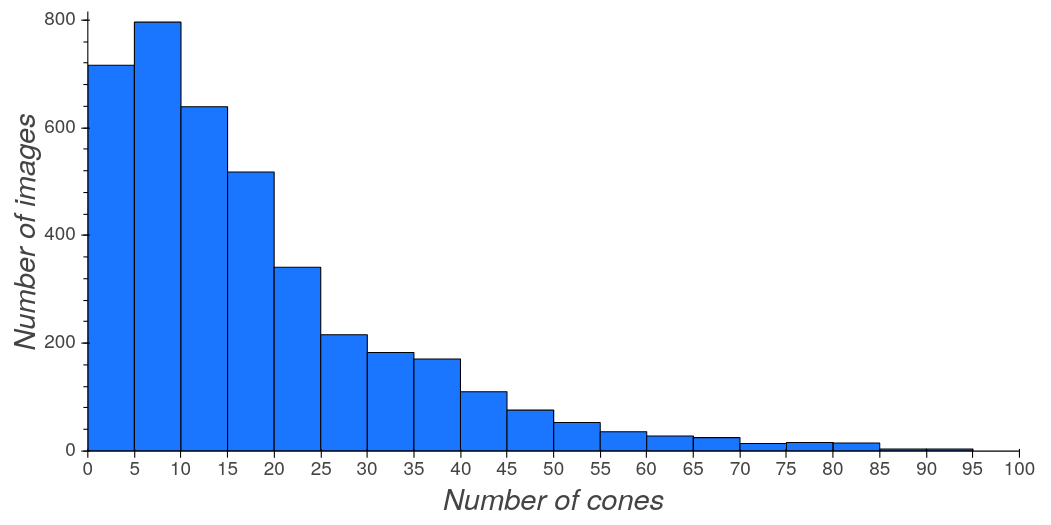}
        \caption{Number of objects per image.}
        \label{sfig:objects-per-img}
    \end{subfigure}
    \\[1ex]
    \begin{subfigure}{0.495\linewidth}
        \centering
        \includegraphics[width=\linewidth]{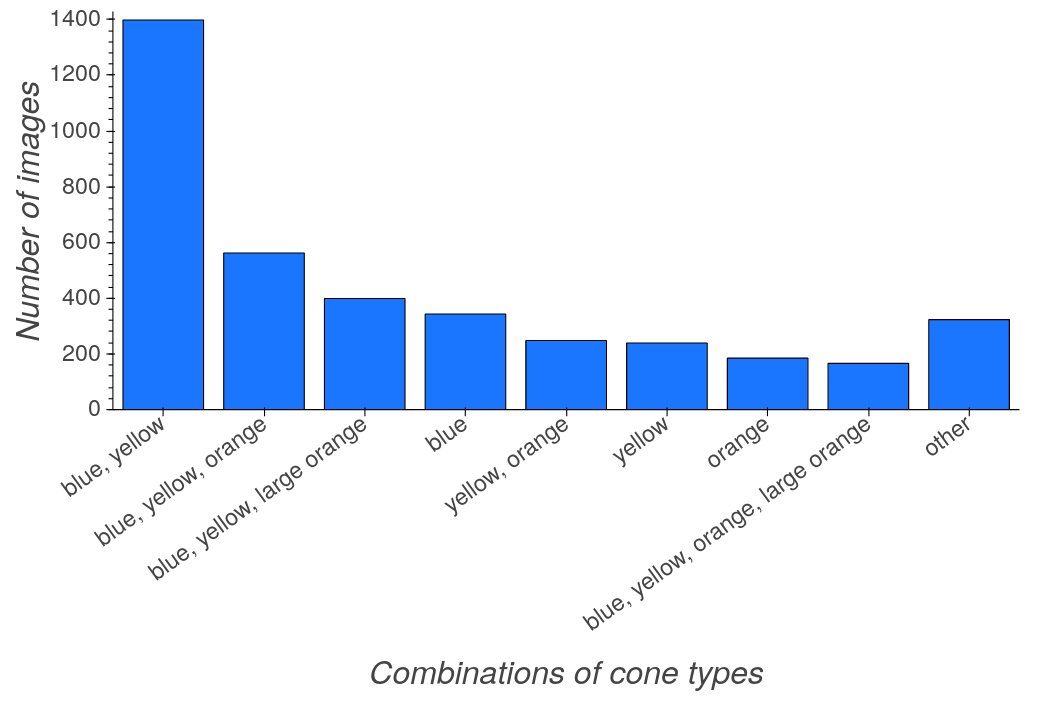}
        \caption{Object class combinations per image. \textit{Other} aggregates the remaining unique combinations.} 
        \label{sfig:scene-comp}
    \end{subfigure}
    \begin{subfigure}{0.495\linewidth}
        \centering
        \includegraphics[width=\linewidth]{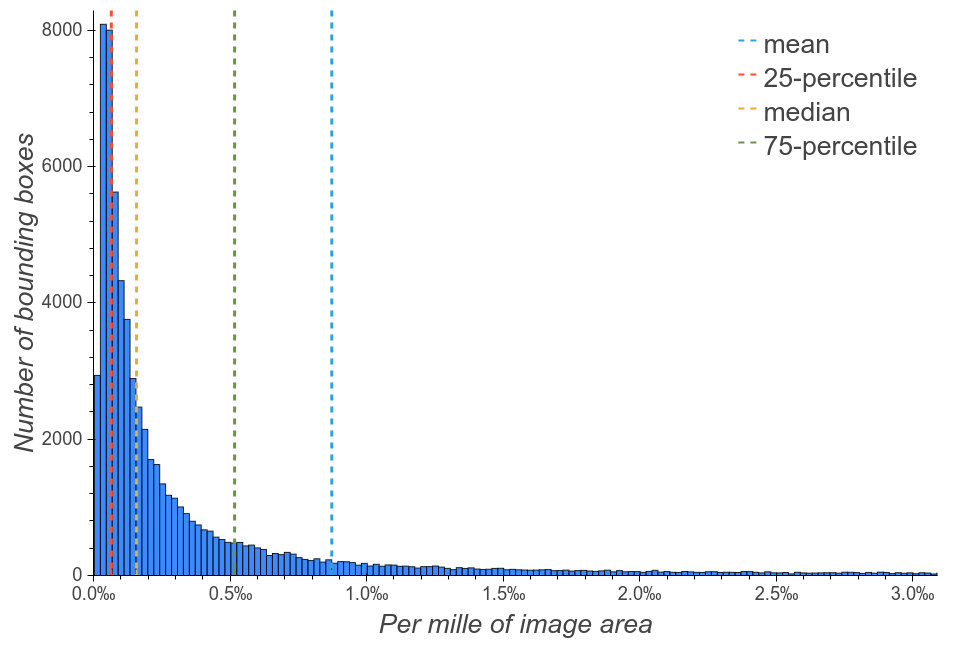}
        \caption{Relative size of bounding boxes with respect to image area.}
        \label{sfig:obj-size}
    \end{subfigure}
    \caption{A quantitative analysis of the dataset's scene complexity shows that there is a small number of classes and distinct combinations thereof. On average, images further contain a large number of very small objects. Together, these characteristics point to classification being relatively easy while precise predictions being difficult.}
    \label{fig:dataset_stats}
\end{figure*}

To analyze the complexity of scenes in the dataset, we quantitatively assess characteristics of their contained instances. From \reffig{sfig:classes-per-img},~\ref{sfig:scene-comp} we can see a certain bias towards two-class scenes, more specifically towards scenes that contain blue and yellow cones, as expected from the rules mentioned in \refsec{sec:introduction}. Note that there are no images with zero distinct cone types since every image in the dataset contains objects.
Moreover, only seven of the 25 possible class combinations are represented in more than 200 images, which points to predictable scene compositions. This constrained predictability is expected due to the way the tracks are built.

As seen in \reffig{sfig:objects-per-img}, the number of instances per image varies greatly. Additionally, \reffig{sfig:obj-size} shows that most instances are of very small size. Together, this poses a challenge to the used object detection algorithms, as very small objects have to be detected in large numbers.

\section{Data Collection}
\label{sec:data_collection}

FSOCO legacy had no guidelines for what kind of raw data to use for contributions. This permissive policy lead to the dataset containing a large portion of very similar images. After performing qualitative analysis, the two most common causes for this were images sourced from one video stream in close succession, and images from the same race track.
To elaborate on the latter, since test days are costly for teams concerning their resources, they are seldom. Furthermore, building rules-compliant tracks takes a considerable amount of time, even for well-practiced teams, i.e., the number of different race tracks per test day is minimal. Compounding this, most teams only have access to a single testing site, leading to very similar background settings even in raw data collected on different days and tracks.

In supervised learning settings, we wish to train models that can generalize to unknown data. In the context of \ac{FSD}, models have to detect cones reliably on any possible track and under any weather conditions.
To accomplish this, the underlying dataset must be representative of the target distribution. Models can also work with data significantly different from the training set, which is referred to as \ac{OOD}. Inference on \ac{OOD} inputs performs much worse than on in-distribution inputs and is a focus of ongoing research~\cite{Ren2019}.
There are approaches like domain adaptation that circumvent needing datasets for the target distribution, which come with considerable additional effort~\cite{Hsu2020}.
Therefore, we focus on designing a training dataset representing the \ac{FSD} competitions.

\subsection{Contribution Procedure}
\label{ssec:contribution_procedure}
The contribution procedure for the FSOCO dataset is the process from the moment a team's contribution request has been accepted to their contribution being approved.
To explain this process, we need to keep the two contribution types to the FSOCO dataset in mind: 1) dataset contributions and 2) label contributions. Dataset contributions entail raw data and labels, while label contributions only include labels of raw data previously donated by other teams. 

Upon being accepted, the team needs to pass a labeling exam consisting of a small raw data sample and perfect ground truth labels that are not visible to the examinee. They must label the raw data following the labeling guidelines and receive feedback based on an automated comparison to the ground truth.
The set of images used in the exam is designed to cover difficult scenes and provides examples for all guideline rules.

After passing the exam, teams either have their dataset reviewed, in the case of a dataset contribution, or are assigned a pre-labeled donated dataset for label contributions. Pre-labels are generated by us using the model introduced in \refsec{ssec:bounding_box_regression}. By leveraging predictions of current models, there are two main advantages. First, since annotating mostly consists of improving existing labels, the achieved quality increases. The adjustment of pre-labels can be seen as reviewing the model's output. 
Second, the human annotator can label objects missed by the model, thereby shrinking the set of \ac{OOD} data.

The first review step is passing automated label sanity checks. These are simple rule-based tests ensuring basic quality measures, e.g., identifying tiny bounding boxes created by mistake.
What follows is an iterative review process, performed by domain-knowledge specialists and based on the labeling guidelines to ensure the highest quality.
Each image is either passed or rejected, and image-level or object-level feedback is given in the case of a rejection. As soon as all images have passed, the contribution is accepted and added to the FSOCO dataset. 

\subsection{Candidate Images}
\label{ssec:candidate_images}
\begin{figure}[t]
    \centering
    \captionsetup[subfigure]{justification=centering}
    \begin{subfigure}{0.495\linewidth}
        \centering
        \includegraphics[width=\linewidth]{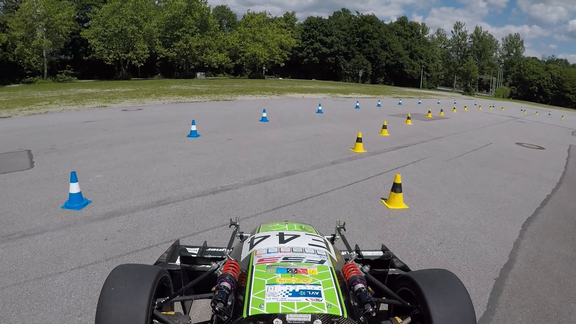}
        \caption{\(K_{cos}(\mathbf{x_a},\mathbf{x_a}) = 1.0\)}
    \end{subfigure}
    \begin{subfigure}{0.495\linewidth}
        \centering
        \includegraphics[width=\linewidth]{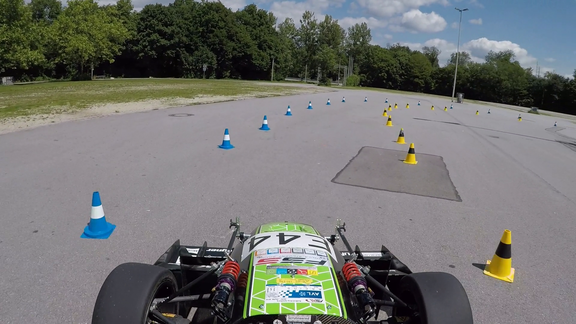}
        \caption{\(K_{cos}(\mathbf{x_a},\mathbf{x_b}) = .990\)}
    \end{subfigure}
    \\[1ex]
    \begin{subfigure}{0.495\linewidth}
        \centering
        \includegraphics[width=\linewidth]{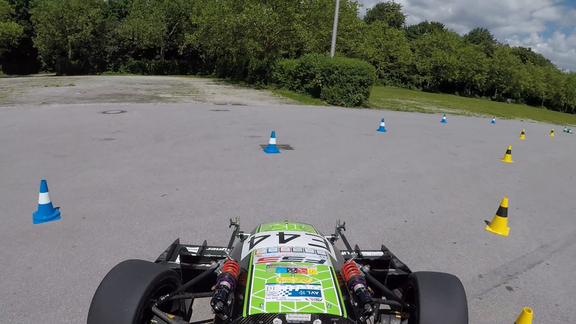}
        \caption{\(K_{cos}(\mathbf{x_a},\mathbf{x_c}) = .985\)}
    \end{subfigure}
    \begin{subfigure}{0.495\linewidth}
        \centering
        \includegraphics[width=\linewidth]{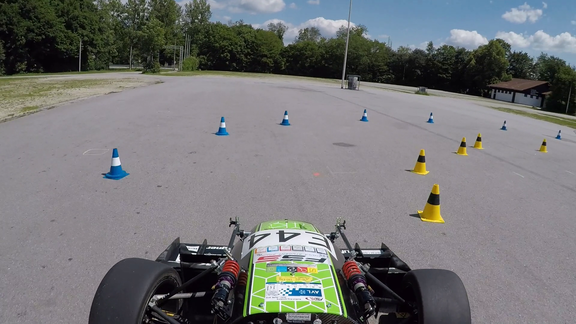}
        \caption{\(K_{cos}(\mathbf{x_a},\mathbf{x_d}) = .981\)}
    \end{subfigure}
    \caption{The image similarity scorer allows to sensibly select a set of images for manual labeling. The given scores are the cosine similarities with respect to the top left image.}
    \label{fig:image_similarity_scorer}
\end{figure}

As previously mentioned, our dataset is designed to enable supervised learning models to generalize to \ac{FSD} competitions. This entails building a training set representative of those settings.
To this end, we employ a two-stage data selection process. First, any contributed dataset is required to consist of at least 50\%  on-board footage of rules-compliant tracks. An example of such footage can be seen in \reffig{fig:image_similarity_scorer}. Second, the submitted datasets need to have a low enough similarity score.

The former measure ensures the collected data is close enough to the target distribution. The latter enforces diversity by only allowing sufficiently dissimilar images.
Additionally, the similarity scorer presented in \refsec{ssec:image_similarity_scorer}, with its automated selection feature, lends itself well to the \ac{FSD} setting. Teams have much recorded on-board data but often lack the knowledge of how to source raw data appropriate for machine learning models properly. The similarity scorer aids in this regard, sampling optimally diverse raw data from the team's raw data source. 
Diversity across different datasets is not tested for since most teams have access to different testing grounds.

\section{Toolset}
\label{sec:toolset}
The FSOCO Tools \ac{CLI} facilitates preparing a contributor's dataset to be added to the project.
It is a small, extensible tool for data manipulation and analysis that is based on Click\footnote{\tt\small\href{https://click.palletsprojects.com/en/7.x/}{click.palletsprojects.com}} to combine scripts into a uniform package. The source code is available in our public GitHub repository\footnote{\tt\small\href{https://github.com/fsoco/fsoco-devkit}{github.com/fsoco/fsoco-devkit}} and can be easily extended.

\subsection{Image Similarity Scorer}
\label{ssec:image_similarity_scorer}
As reasoned in \refsec{ssec:candidate_images}, candidate images are selected based on their relative visual similarity. On this end, the developed image similarity scorer assigns a grade to each image encoding its likeness with respect to all other images in the processed input folder.

First, utilizing a publicly available version of the AlexNet~\cite{Krizhevsky2012} convolutional neural network, pre-trained on ImageNet~\cite{Deng2009}, feature maps are extracted for every image. In detail, feature vectors of length 4,096 are obtained from the 3rd convolutional layer. This layer has been proven to be a good balance between strong pixel sensitivity in early layers and very general abstractions in higher layers. Second, the scorer computes the pairwise cosine similarity between all image pairs using the feature maps.

The following describes the calculation of the similarity score. Let $F$ be the set of all extracted image feature vectors of the specific dataset consisting of $n$ images. The cosine similarity of $\mathbf{x}, \mathbf{y} \in F$ is then given by \refeqn{eqn:1}. Note that a similarity of 1 implies $\mathbf{x}$ and $\mathbf{y}$ being the same vector, while -1 denotes exact opposites.
\\[-.5pt]
\begin{equation}
    \label{eqn:1}
    K_{cos}(\mathbf{x},\mathbf{y}) = \frac{\mathbf{x} \cdot \mathbf{y}}{||\mathbf{x}|| \ ||\mathbf{y}||}
\end{equation}
\\[-.5pt]
The 95\% similarity score of the dataset's $k^{th}$ image with feature vector $\mathbf{x_k}$ is computed as
\\[-.5pt]
\begin{equation}
    \label{eqn:2}
    image\_score_{0.95}(\mathbf{x_k}) = \sum_{\substack{i=1 \\ i\neq k}}^{n} \begin{dcases} 1,& \text{if } K_{cos}(\mathbf{x_k},\mathbf{x_i})\geq 0.95\\0, & \text{otherwise}  \end{dcases} 
\end{equation}
\\[-.5pt]
Finally, the 95\% similarity score of the entire dataset is defined as the average of the individual image scores, i.e., the result of \refeqn{eqn:3} reports the number of similarly looking images a the dataset.
\\[-.5pt]
\begin{equation}
    \label{eqn:3}
    dataset\_score_{0.95} = \frac{1}{n} \sum_{k=1}^{n} image\_score_{0.95}(\mathbf{x_k})
\end{equation}

\subsection{Label Converters}
\label{ssec:label_converters}
Many datasets introduce their own annotation formats adapted to the respective requirements. Additionally, no standard annotation format has been established among the publicly accessible labeling tools. The resulting heterogeneous annotation formats can make aggregating data cumbersome. To accommodate these diverse sources and to ease the contribution process for interested parties, we offer multiple label converters for bounding boxes. Note that the FSOCO dataset contains annotations in the Supervisely format.
\[
\begin{alignedat}{1}
\text{{DarknetYOLO}}\rightarrow & \text{{\,Supervisely}}\\
\text{{LabelBox}}\rightarrow & \text{{\,Supervisely}}\\
\text{{Supervisely}}\rightarrow & \text{{\,DarknetYOLO}}\\
\text{{Supervisely}}\rightarrow & \text{{\,Pascal\,VOC}}
\end{alignedat}
\]

The formats listed above have been prioritized for implementation as they are being widely used in the \ac{FSD} community. In particular, a converter to Pascal VOC is available for compatibility with already existing label converters to other formats. Both decisions simplify contribution to and usage of the dataset.

\subsection{Miscellaneous Tools}
\label{ssec:miscellaneous_tools}

The FSOCO Tools \ac{CLI} includes an easy-to-use data visualizer to facilitate inspection of the dataset without further processing. It supports bounding boxes in both Supervisely and Darknet YOLO format and segmentation labels from Supervisely.

In order to give credits to the contributing teams, a watermark is added to all images in the dataset, consisting of the team's logo and a timestamp. When processing the images, e.g., training a neural network, the watermark should be removed by cropping the outer 140 pixels on all image borders.

\section{Tasks}
\label{sec:tasks}
To evaluate our selection procedure for candidate images, we compare the deprecated FSOCO legacy version with the newly released FSOCO dataset. We further provide a baseline for the bounding box regression task to demonstrate the possible performance when leveraging FSOCO for cone detection.

\subsection{Candidate Images Selection}
\label{ssec:candidate_images_selection}
We analyze the efficacy of the candidate image selection method, presented in \refsec{ssec:candidate_images}, by computing two image similarity scores for both FSOCO and FSOCO legacy encoding the image similarities within each teams' contribution as well as across all contributions. 

\begin{figure*}[t]
    \centering
    \captionsetup[subfigure]{justification=centering}
    \begin{subfigure}{0.495\linewidth}
        \centering
        \includegraphics[width=\linewidth]{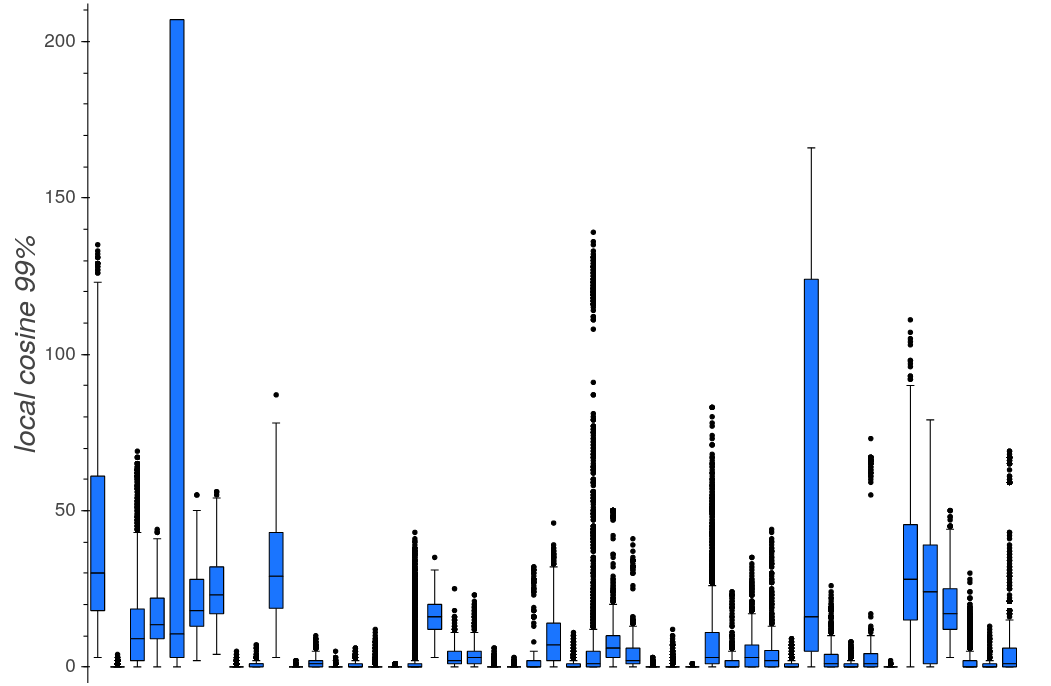}
        \caption{Deprecated version: FSOCO legacy}
    \end{subfigure}
    \begin{subfigure}{0.495\linewidth}
        \centering
        \includegraphics[width=\linewidth]{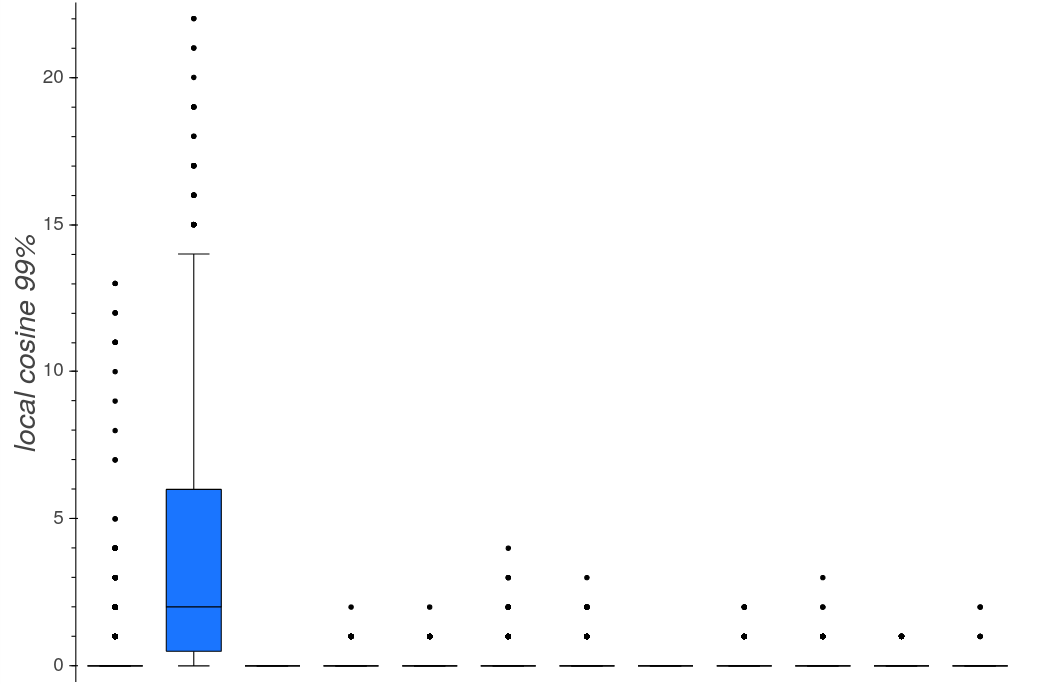}
        \caption{The FSOCO dataset}
    \end{subfigure}
    \caption{Comparison of 99\% cosine similarity scores within the respective datasets. Team names are omitted on purpose to preserve their privacy. Note that the allowed range of similarity scores varies depending on the type of data, i.e., official FSD event versus testing tracks and off-track data.}
    \label{fig:local_cosine_box_plots}
\end{figure*}

The local scores, i.e., separated by contributors, are depicted in \reffig{fig:local_cosine_box_plots} showing the apparent differences between the legacy version and our dataset. The higher the cosine score, the more similar images the reference set contains. For instance, a score of 5.0 indicates that, on average, for each image in the set there are five images with a cosine similarity of at least 99\% based on the extracted feature vectors. Taking into account that \reffig{fig:local_cosine_box_plots} shows the cosine 99\% score, the reported numbers of duplicates should be interpreted as conservative estimates.

\begin{table}[h]
    \centering
    \fontsize{8}{10}\selectfont
    \caption{Comparison of cosine similarity scores across the entire datasets. The values indicate the average number of similar images for different similarities.}
    \label{tab:global_cosine_similarity}
    \setlength{\tabcolsep}{5.5pt}
    \begin{tabular}{lccc}
    \toprule
     & \makecell{global cosine \\ 99\%} & \makecell{global cosine \\ 98\%} & \makecell{global cosine \\ 95\%} \\
    \midrule
    FSOCO legacy & 8.97 & 28.85 & 163.65 \\
    FSOCO & 1.15 & 8.63 & 160.44 \\
    \bottomrule
    \end{tabular}
    \vspace{1ex}
\end{table}

\reftab{tab:global_cosine_similarity} presents global cosine similarities for 95\%, 98\%, and 99\% levels of both FSOCO and FSOCO legacy, i.e., the metrics are based on the contributions of all teams. Similar to the local scores, our dataset has significantly lower values than the legacy version. Intuitively speaking, many of the images in FSOCO legacy might not add new information to a learning-based cone detection algorithm and, thus, should be discarded to reduce the required training time. Note that we do not impose requirements for the global similarity score and yet, by enforcing small local values and diverse contributions from different raw data sources, we are able to also achieve low global scores.

\subsection{Bounding Box Regression}
\label{ssec:bounding_box_regression}

\begin{table}[t]
    \centering
    \fontsize{8}{10}\selectfont
    \caption{Overview of all datasets used for either training or testing.}
    \label{tab:dataset_stats}
    \setlength{\tabcolsep}{5.5pt}
    \begin{tabular}{lccc}
    \toprule
    & \makecell{Number of\\images} & \makecell{Number of\\cones} & \makecell{Cones per\\image}\\
    \midrule
    FSOCO legacy & 51,047 & 394,728 & 7.73 \\
    FSOCO & 3,821 & 64,905 & 17.77 \\
    \midrule
    FSOCO legacy test set & 963 & 18,727 & 19.45 \\
    FSOCO test set & 250 & 3,704 & 14.82 \\
    \bottomrule
    \end{tabular}
    \vspace{1ex}
\end{table}

We compare the performance of the FSOCO dataset with the deprecated legacy version using YOLOv4~\cite{Bochkovskiy2020} as a state-of-the-art object detection network. Both training and inference are done utilizing the authors' publicly available codebase applying the default settings and a batch size of 64 with subdivision set to 64. All experiments are conducted on Nvidia Titan V and 2020 Ti GPUs. Regardless of the size of the dataset, we train for 10,000 batches, i.e., in smaller datasets the same image is seen more often than in larger datasets. The best test accuracy was mostly reached between the 4,000\textsuperscript{th} and 6,000\textsuperscript{th} batch. Therefore, we assume that this training period is sufficiently long.

Average precision values for several intersection of union thresholds are reported for two different test sets: 1) a non-public FSOCO test set and 2) the FSOCO legacy version of a single contributor, whose quality has been manually verified. Note that there is no overlap between any training and testing sets. Furthermore, the FSOCO legacy test set contains very few images from the testing sites covered in the FSOCO dataset. A complete overview of all utilized datasets is given in \reftab{tab:dataset_stats}. As the legacy version does not have standardized object classes, we focus on the regression task. However, tests have shown that the classification error on the four classes is below 5\% and, thus, not the decisive factor.

\begin{figure}[b]
    \vspace{2ex}
    \centering
    \includegraphics[width=\linewidth]{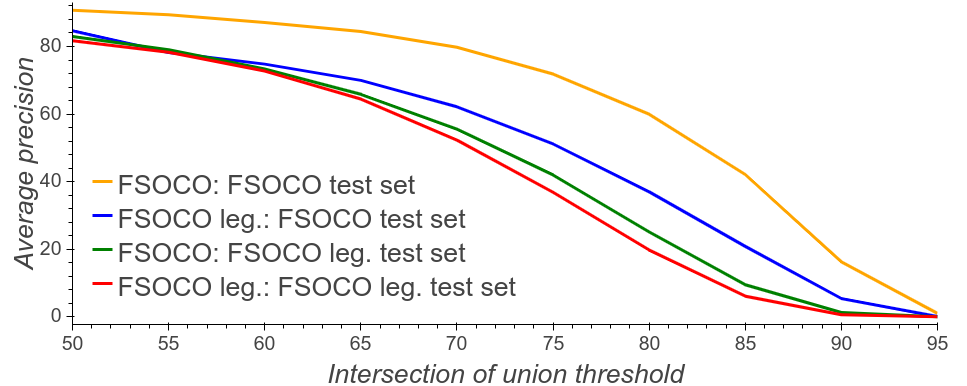}
    \caption{Comparison of the average precision of YOLOv4 trained on several FSOCO subsets on a test set extracted from FSOCO legacy and the non-public FSOCO test set.}
    \label{fig:average_precision}
\end{figure}

As reported in \reffig{fig:average_precision}, training with data only from our new approach of FSOCO outperforms the unpolished legacy version on both test sets. Moreover, with increasing the required intersection of union for a prediction to count towards the average precision, the difference in performance between both approaches increases until the task cannot be sensibly learned by the employed model. The improved performance for more precise bounding boxes is expected as the quality requirements are higher for FSOCO, see~\refsec{sec:data_collection}.
Note that the FSOCO dataset comprises just a fraction of the number of images with respect to its predecessor. In particular, at the time of submission FSOCO contains only 7.5\% and 16.4\% of the number of images and cone annotations compared to FSOCO legacy.

\section{Summary and Future Work}
\label{sec:summary}
This paper presented FSOCO, a collaborative data buy-in dataset for camera-based cone detection systems for \ac{FSD}. Leveraging a simple toolset and data collection strategy, it enables interested parties to work together towards building a dataset for supervised learning tasks of high quality. In a comparative analysis we show a baseline for bounding box regression to perform better than on a previous version of this dataset while comprising less annotations.

Currently the main supported task is bounding box regression, although an instance segmentation mask dataset is available. In the future, we plan to expand the dataset with further data modalities, e.g., LiDAR point clouds.
Finally, a small part of each contribution is added to a non-public test dataset. This test dataset can help enable task-specific machine learning competitions, similar to the ImageNet challenges, for \ac{FSD} in the future.

\addtolength{\textheight}{-12cm}   


\section*{Acknowledgments}

FSOCO is only possible due to the numerous contributions of the \mbox{Formula} Student Driverless community. We would like to thank all participating teams for helping us to accelerate the development of autonomous racing.
We further thank the Autonomous Racing Workshop for providing a discussion platform for interested student teams leading to this project.



\bibliographystyle{IEEEtran}
\bibliography{Bibliography}

\end{document}